\documentclass{article}
\usepackage[numbers]{natbib}
\usepackage[final]{nips_2017} 
\usepackage[utf8]{inputenc} 
\usepackage[T1]{fontenc}    
\usepackage{hyperref}       
\usepackage{url}            
\usepackage{booktabs}       
\usepackage{amssymb}
\usepackage{amsmath}
\usepackage{xcolor}
\usepackage{caption}
\usepackage{amsfonts}       
\usepackage{nicefrac}       
\usepackage{microtype}      
\usepackage{algorithmic, algorithm}	
\usepackage{graphicx}
\def\thetabold{\mathbf{\theta}}
\def\vbold{\mathbf{v}}
\newcommand{\multiauthor}[1]{#1 \textit{et al}}

\title{\textit{ADINE}: An Adaptive Momentum Method for Stochastic Gradient Descent}
\author{
Vishwak Srinivasan \hspace{7.5mm} \hfill Adepu Ravi Sankar \hspace{7.5mm} \hfill Vineeth N Balasubramanian\\
Indian Institute of Technology Hyderabad\\
\texttt{\{cs15btech11043, cs14resch11001, vineethnb\}@iith.ac.in}
}
\begin{document}
\maketitle
\begin{abstract}
Two major momentum-based techniques that have achieved tremendous success in optimization are Polyak's heavy ball method and Nesterov's accelerated gradient. A crucial step in all momentum-based methods is the choice of the momentum parameter $m$ which is always suggested to be set to less than $1$. Although the choice of $m < 1$ is justified only under very strong theoretical assumptions, it works well in practice even when the assumptions do not necessarily hold. In this paper, we propose a new momentum based method \textit{ADINE}, which relaxes the constraint of $m < 1$ and allows the learning algorithm to use adaptive higher momentum (also called `inertia', hence the name \textit{ADINE}). We motivate our hypothesis on $m$ by experimentally verifying that a higher momentum ($\ge 1$) can help escape saddles much faster. Using this motivation, we propose our method \textit{ADINE} that helps weigh the previous updates more (by setting the momentum parameter $> 1$), evaluate our proposed algorithm on deep neural networks and show that \textit{ADINE} helps the learning algorithm to converge much faster without compromising on the generalization error.
\end{abstract}

\section{Introduction}
\vspace{-8pt}
Non-convex optimization problems are natural formulations in many machine learning problems (e.g. (Un)supervised learning, Bayesian learning). Various learning approaches have been proposed in such settings, as global minimization of such problems are NP-hard in general. Gradient descent is the de-facto iterative learning algorithm used for such optimization problems in machine learning, especially in deep learning. Several variants of gradient descent methods have been proposed and all thse proposed methods can be broadly classified into momentum-based methods (e.g. \textit{Nesterov's Accelerated Gradient} \cite{nesterov1983method}), variance reduction methods (e.g. \textit{Stochastic Variance Reduced Gradient \cite{Johnson:2013:ASG:2999611.2999647},\cite{reddi2016stochastic}}) and adaptive learning methods (e.g. \textit{AdaGrad} \cite{Duchi:EECS-2010-24}). 

Gradient descent coupled with momentum - also called \emph{classical momentum} by Polyak \cite{polyak1964some}, is the first ever variant of gradient descent involving the usage of a momentum parameter. The momentum methods use the information from previous gradients in addition to the current gradient for updating the learning parameters. Nesterov in his seminal work \cite{nesterov1983method}, proposed an accelerated gradient method (also a momentum based method as shown by \cite{Sutskever:2013:IIM:3042817.3043064}) which gives an upper bound on the number of iterations for learning algorithm to converge. With the tremendous success of deep learning models,
\multiauthor{Sutskever} in their work \cite{Sutskever:2013:IIM:3042817.3043064} worked out to incorporate the algorithm by Nesterov \cite{nesterov1983method}. Nesterov's method performs an update in the same way as \textit{classical momentum}, only with a correction to the gradient.

Gradient descent is generally used in the form of \textit{mini-batch gradient descent} in minimizing all real world optimization problems, where only a small subset of training data (called a mini-batch) is used due to the presence of enormous amounts of training data. The use of a mini-batch for gradient calculation introduces a lot of variance due to the stochasticity of learning algorithm. Methods like SVRG \cite{Johnson:2013:ASG:2999611.2999647}, \cite{reddi2016stochastic} have been proposed, which try to reduce the variance in gradient with strong theoretical guarantees. There exist other variance reduction methods like SAG \cite{roux2012stochastic} and SDCA \cite{shalev2013stochastic} also. 

Recently, several methods have been proposed that try to adapt the learning rate in gradient descent. Riedmiller and Braun proposed \emph{Rprop} \cite{298623} method which suggested the usage of an adaptive learning rate based on the sign of gradient in last two iterates. \emph{Rprop} increases the learning rate of a weight if the gradient sign does not change in last two iterates, otherwise it decreases the learning rate. \emph{AdaGrad} - proposed by \multiauthor{Duchi} \cite{Duchi:EECS-2010-24}, divides $\eta$(a global learning rate) of every step by the square of the $\ell_2$ norm of all previous gradients. This scaling using the norm reduces the learning in dimensions which have already changed significantly, and speeds up in the dimensions that have not changed rapidly, thereby stabilizing the model. \emph{RMSProp} proposed by \multiauthor{Tieleman} \cite{Tieleman2012} is a simple amalgamation of \emph{Rprop} and \emph{SGD}. This method scales the learning rate by the decaying average of squared gradient. There are few other methods proposed which adapts the learning rate like AdaDelta \cite{zeiler2012adadelta}. \emph{Adam} - proposed by Kingma and Ba \cite{kingma2014adam} is a very successful method that almost all recent state-of-the-art deep learning models used. \emph{Adam} makes use of the first and second order moments of gradients and ideas from norm-based methods, through combining the advantages from AdaGrad and RMSProp.

In this work, we study momentum-based methods and propose the idea of having scheduled increased momentum. We motivate our work by showing that higher momentum can help escape saddles. We then propose \textit{ADINE} - an adaptive momentum based method which helps learning algorithms converge faster. The paper is organized as follows: Section \ref{prev-work} discusses existing methods and background, Section \ref{motivation} discusses the motivation of our work and shows how higher momentums help escape saddles, Section \ref{algorithm} showcases our proposed algorithm \textit{ADINE} and Section \ref{experiment} contain experimental results that validate our hypothesis.

\section{Background and Previous Work}
\label{prev-work}
\vspace{-8pt}
Let us consider the minimization of a function \(f\) w.r.t. parameters denoted by \(\thetabold\). If this function is convex in nature, the minimization is "easy". But in most real-world problems, such as in deep learning, this function is non-convex in nature, making it "difficult". The \textit{gradient descent (abbrev. GD)} algorithm has become the workhorse to minimize such non-convex functions. In 1964, Polyak \cite{polyak1964some} proposed a way to use the previous updates in \textit{GD}, and this is referred to as \textit{Polyak's heavy ball method} or \textit{Classical Momentum (abbrev. CM)}. The update equations of \textit{CM} are:
\begin{gather}
\label{polyak-1}
\vbold_{t+1} = m\vbold_{t} - \eta \nabla f\left(\thetabold_{t}\right) \hspace{5mm} \text{with} \hspace{1mm} \vbold_{0} = \mathbf{0}\\
\thetabold_{t+1} = \theta_{t} + \vbold_{t+1}
\end{gather}

Here, \(m\) is the momentum parameter, and \(\eta\) is the learning rate. However, this method is proven to be theoretically beneficial under very strong conditions (strong convexity and strong smoothness). In 1983, \textit{Nesterov's Accelerated Gradient (abbrev. NAG)} \cite{nesterov1983method} given by Nesterov, achieved convergence at the rate of \(O\left(\nicefrac{1}{k^2}\right)\) under minimal assumptions (Lipschitz continuity). The update equations of \textit{NAG} are given below, where \(m_{t}\) and \(\eta\) stand for momentum and learning rate respectively.
\begin{gather}
\mathbf{y}_{t+1} = \thetabold_{t+1} + m_{t}\left(\thetabold_{t} - \thetabold_{t-1}\right) \hspace{5mm} \text{with} \hspace{1mm} \thetabold_{0} = \thetabold_{-1}\\
\thetabold_{t+1} = \mathbf{y}_{t+1} - \eta \nabla f\left(\mathbf{y}_{t+1}\right) \\
m_{t} = \frac{a_{t} - 1}{a_{t+1}} \\
a_{t} = \frac{1 + \sqrt{4a_{t-1}^2 + 1}}{2} \text{ and } a_{0} = 1
\end{gather}

In our work, we consider the analogous update equations presented by \multiauthor{Sutskever} \cite{Sutskever:2013:IIM:3042817.3043064} which characterize \textit{NAG}. These are as follows:
\begin{gather}
\vbold_{t+1} = m\vbold_{t} - \eta \nabla f\left(\thetabold_{t} + m\vbold_{t}\right) \hspace{5mm} \text{with} \hspace{1mm} \vbold_{0} = \mathbf{0}\\
\thetabold_{t+1} = \thetabold_{t} + \vbold_{t+1}
\end{gather}

\subsection*{Theoretical guarantees from \textit{CM}}
Let \(f(\cdot)\) be an \(\alpha\)-strongly convex and \(\beta\)-smooth function on \(\mathbb{R}^{d}\). The heavy ball method achieves an \(\epsilon\)-accurate solution of \(\thetabold^{*}\) in
\begin{equation*}
O\left(\sqrt{\kappa}\log\left(\frac{1}{\epsilon}\right)\right)
\end{equation*}
steps, where \(\kappa := \frac{\beta}{\alpha}\) is the condition number of the problem and \(\eta_{\text{opt}}\) and \(m_{opt}\) in equation \ref{polyak-1} is given by 
\begin{equation*}
\eta_{\text{opt}} = \frac{4}{(\sqrt{\beta} + \sqrt{\alpha})^2} \quad, \quad m_{\text{opt}} = \left(\frac{\sqrt{\beta} - \sqrt{\alpha}}{\sqrt{\beta} + \sqrt{\alpha}}\right)^2
\end{equation*}

\subsection*{Theoretical guarantees from \textit{NAG}}
Let \(f(\cdot)\) be a convex function over \(\mathbb{R}^{d}\) and \(L\)-smooth, then \textit{NAG} achieves an \(\epsilon\)-accurate solution of \(\thetabold^{*}\) in
\begin{equation}
O\left(\sqrt{\frac{L}{\epsilon}}\right)
\end{equation}
steps, and also ensures that:
\begin{equation}
f(\thetabold^{(t)}) - f(\thetabold^{*}) = O\left(\frac{1}{t^2}\right)
\end{equation}

\section{Motivation}
\label{motivation}
\vspace{-8pt}
A key motivation for our work arises from observing the values of the momentum parameters in \textit{CM} and \textit{NAG}. Polyak's suggested choice of \(m < 1\) in his method is a general scenario while making use of \textit{CM} in most optimization problems. We analyze the reason for the choice of \(m < 1\) in \textit{NAG}. We refer to the analogue to \(a_{t}\) used by \multiauthor{Sutskever} in \cite{Sutskever:2013:IIM:3042817.3043064}, and its consequence in the approximation of the momentum parameter \(m_{t}\), which is:
\begin{gather}
a_{t} = \frac{1 + \sqrt{4a_{t-1}^{2} + 1}}{2} \approx a_{t-1} + 0.5 \hspace{5mm} \left(\text{For large } a_{t}\right) \\
\implies a_{t} \approx \frac{t+4}{2} \implies m_{t} = \frac{a_{t} - 1}{a_{t+1}} \approx \frac{t+2}{t+5} < 1
\end{gather}

We take \textit{CM} and analyze the behaviour of the update focusing on the importance of the momentum parameter. We perform a sum over the parameter updates in \textit{CM}, with \(f(\cdot)\) in consideration:
\begin{equation}
\setlength{\abovedisplayskip}{3pt}
\setlength{\belowdisplayskip}{3pt}
\displaystyle \thetabold_{t+1} = \thetabold_{t} - \eta \nabla f\left(\thetabold_{t}\right) + m(\thetabold_{t} - \thetabold_{t-1}) \implies \thetabold_{n+1} = m\thetabold_{n} + (1 - m)\thetabold_{0} - \eta \sum_{t=0}^{n} \nabla f\left(\thetabold_{t}\right)
\end{equation}

Note the effect of \(m\) on the previous update, which suggests that higher the momentum parameter, higher the influence of the previous update over the update. Such a correspondence can also be drawn with respect to \textit{NAG} as well.

The study of momentum in  a non-convex setup is interesting because the loss function of a deep neural network (objective function) to be optimized doesn't satisfy the convexity assumptions Polyak and Nesterov make, and yet has achieved huge success in recent years, prior to the arrival of newer adaptive gradient methods. We believe that the analysis of momentum is still not addressed in its entirety. We start our analysis by investigating to see if we can do any better with existing momentum-based methods by weighing the previous updates more. We hypothesize that setting the momentum parameter higher will help the learning algorithm to converge faster. Another motivation to study the variation of \(m\) is the problem of saddles in deep learning models. It has been argued that as the dimensionality of the model increases, existence of local minima is no longer an issue; instead, it is the exponential proliferation of saddle points \cite{dauphin2014identifying}\cite{pmlr-v40-Ge15} which makes optimization in deep learning slow. We hypothesize that setting momentum to \(\geq 1\) can help escape saddles. 

\section{Proposed Method: \textit{ADINE}}
\label{algorithm}
\vspace{-8pt}
As mentioned earlier, in most real-world settings for performing  gradient descent, especially in deep learning, a variant of gradient descent - \textit{mini-batch gradient descent} is used. However, the loss computed over a mini-batch is a noisy estimate of the actual loss and depends on the size of the mini-batch taken. To help smoothen this loss relatively, for the use of \textit{ADINE}, we suggest the use of weighted sum loss, which captures the monotonic nature of the descent as well as the noise that is characteristic of this method. We define this weighted sum loss (\textit{abbrev. WSL}) as:
\begin{equation}
\setlength{\abovedisplayskip}{3pt}
\setlength{\belowdisplayskip}{3pt}
\hat{l}_{k} = \frac{\hat{l}_{k-1} + l_{k}}{2} \hspace{3mm} \text{with} \hspace{3mm} \hat{l}_{1} = \frac{l_{1}}{2} \hspace{5mm} \leftrightarrow \hspace{5mm} \hat{l}_{k} = \displaystyle \sum_{i=1}^{k} \frac{l_{i }}{\left(2^{k - i + 1}\right)}
\end{equation}

Here \(l_{k} = l(\mathbf{\theta}_{k})\) and \(\hat{l}_{k}\) stands for the \textit{WSL} computed after \(k\) iterations. Noticeably, the first form is recursive, and the second form is a closed form expression. Using this definition of \textit{WSL}, we propose our ADaptive INErtia algorithm, \textit{ADINE}, below.

\begin{algorithm}[H]
\caption{\textit{ADINE} (\textit{AD}aptive \textit{INE}rtia), the proposed algorithm}
\begin{algorithmic}[1]
\REQUIRE \(\mathbf{\theta}_{0}\): Initial parameters
\REQUIRE \(l(\cdot)\): Loss Function
\REQUIRE \(\eta\): Learning Rate
\REQUIRE \(m_s, m_g\): Standard momentum (\(< 1\)) and Greater momentum (\(\geq 1\))
\REQUIRE \(\zeta > 0\): ADINE hyperparameter to control tolerance
\STATE \(t \leftarrow 0\), \(\hat{l}_{0} \leftarrow 0\), \(\mathbf{v}_{0} \leftarrow \mathbf{0}\) (Initialize timestep, weighted-sum-loss and velocity vector)
\STATE \(m \leftarrow m_s\) (Initialize the momentum parameter)
\WHILE{until convergence of \(\mathbf{\theta}_t\)}
\STATE \(t \leftarrow t + 1\)
\STATE Compute current loss \(l_{t-1} = l(\mathbf{\theta}_{t-1})\) and weighted-sum-loss \(\hat{l}_{t} = \frac{\hat{l}_{t-1} + l_{t}}{2}\)
\IF{\(\hat{l}_{t} > \zeta \hat{l}_{t-1}\)}
\STATE \(m \leftarrow m_s\)
\ELSE
\STATE \(m \leftarrow m_g\)
\ENDIF
\STATE \(\displaystyle \mathbf{v}_{t} = m\mathbf{v}_{t-1} - \eta\nabla l(\mathbf{\theta}_{t-1} + m\mathbf{v}_{t-1})\)
\STATE \(\displaystyle \mathbf{\theta}_{t} = \mathbf{\theta}_{t-1} + \mathbf{v}_{t}\)
\ENDWHILE
\RETURN \(\theta_{*}\) (Resulting Parameters)
\end{algorithmic}
\end{algorithm}

\(\zeta\) is a new hyperparameter that our algorithm takes as input. The intuition for this hyperparameter is as follows: taking into account the noisy nature of updates arising due to mini-batch gradient descent, if the current \textit{WSL} i.e., \(\hat{l}_{k}\) is greater than the previous \textit{WSL} i.e., \(\hat{l}_{k-1}\) within a limit, decided by the $\zeta$ factor, we set momentum to \(m_g\). If the current \textit{WSL} is lower than the previous \textit{WSL}, we are in terms, since this ensures progress. If the current \textit{WSL} is higher than allowed amount of increase dictated by \(\zeta\), then we set momentum to \(m_s\). Setting higher \(\zeta\) allows for higher upper bound/tolerance from above, which could cause your model to train badly. On the contrary, setting lower \(\zeta\) allows for a strict upper bound/tolerance, which could cause momentum to not being set to \(m_{g}\) at all.

We performed some ablation studies on this new hyperparameter, and we list our observations:
\begin{itemize}
\item For wide and shallow networks, a higher choice of \(\zeta\) is preferred for better results.
\item For narrow and shallow networks, a lower choice of \(\zeta\) is preferred for better results.
\item For deep networks, a lower choice of \(\zeta\) is preferred for better results.
\end{itemize}

\section{Experiments}
\label{experiment}
\subsection{Experiments on Synthetic Functions}
\begin{figure}[H]
\begin{minipage}{0.475\linewidth}
\includegraphics[width=0.95\textwidth]{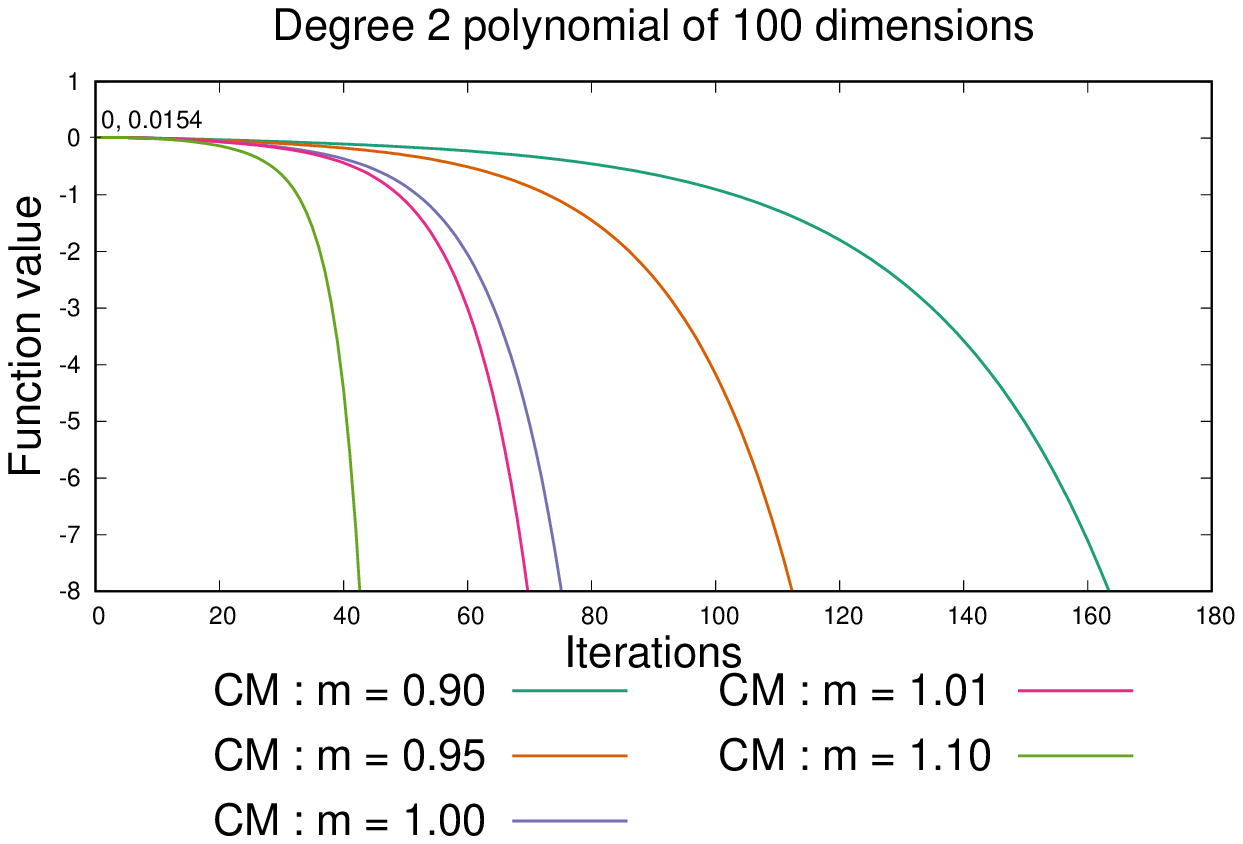}
\end{minipage}
\hfill
\begin{minipage}{0.475\linewidth}
\includegraphics[width=0.95\textwidth]{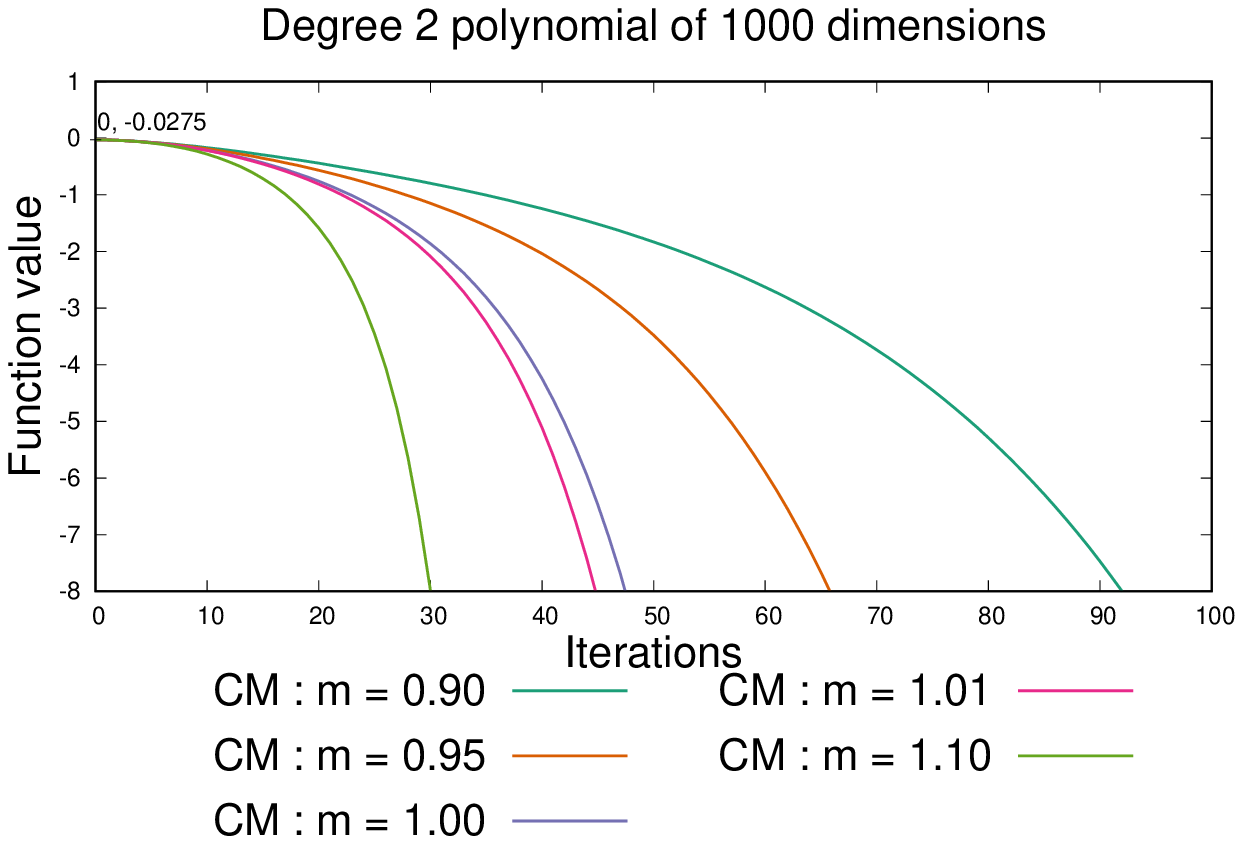}
\end{minipage}

\caption{We set \(n\) to be 100 and 1000 and used \textit{CM} until the value of the function was lesser than -8. We use 5 different momentum parameters from the set \(\{0.90, 0.95, 1.00, 1.01, 1.10\}\). Note how the descent is faster in the case of \(m \geq 1\) Very similar kinds of results were observed with \textit{NAG}. (Best viewed in color)}
\label{fig_synthetic_results-1}
\end{figure}

We first show results to study the hypothesis that higher momentum parameter can help escape saddles. We use a generalized quadratic function of the form
\begin{equation}
\label{gen-quad}
\displaystyle f(\mathbf{x}) = \mathbf{x}^{T} \Lambda \mathbf{x}
\end{equation}
where \(\Lambda\) is a diagonal matrix whose entries are sampled from \(\mathcal{U}[0.99, 1.01]\) and we toggle the signs of the \(\Lambda_{ii}\)'s alternatively, to ensure that the the critical point \(\mathbf{x}^{*} = \mathbf{0}\) is a saddle point. \(n\) is the dimensionality of the domain of the function \(f\). The results are shown in Figure \ref{fig_synthetic_results-1}.

\begin{figure}[H]
\begin{minipage}{0.475\linewidth}
\includegraphics[width=0.95\textwidth]{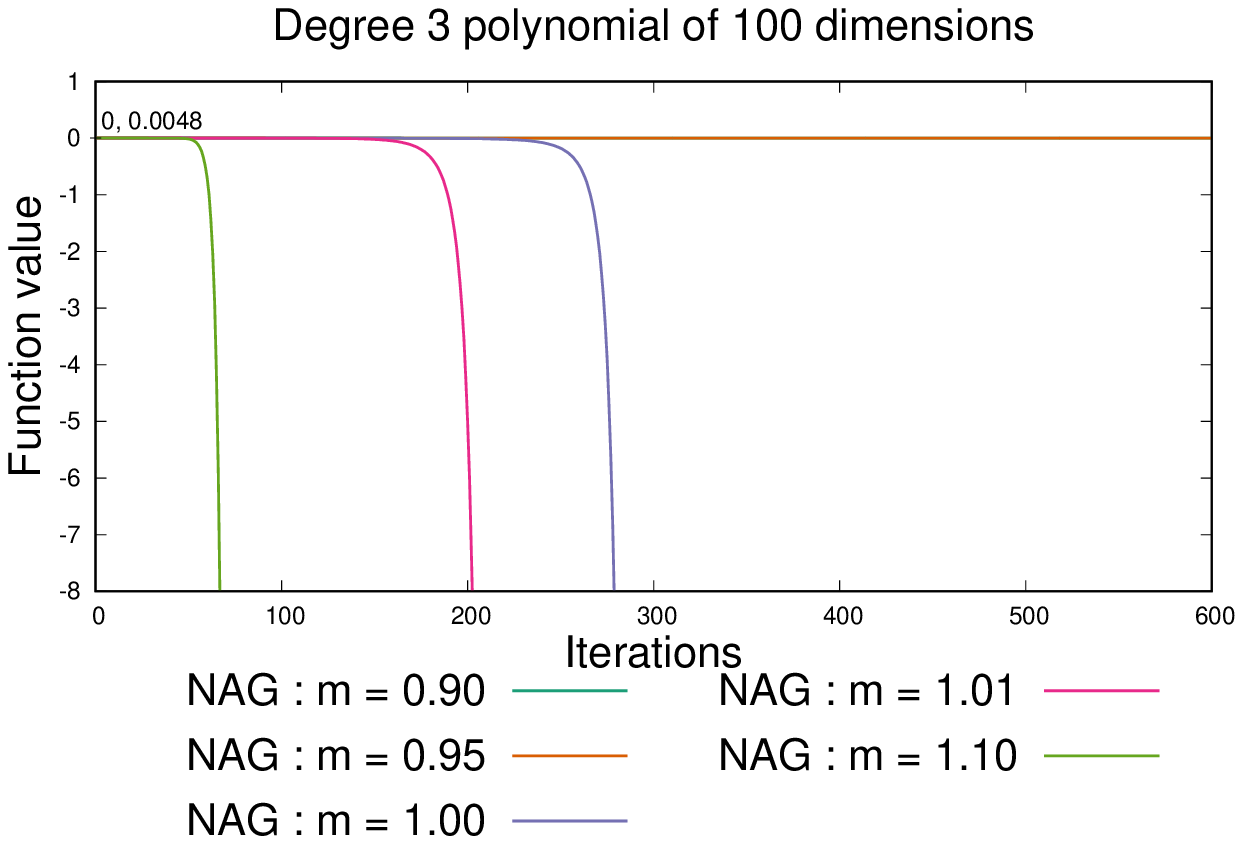}
\end{minipage}
\hfill
\begin{minipage}{0.475\linewidth}
\includegraphics[width=0.95\textwidth]{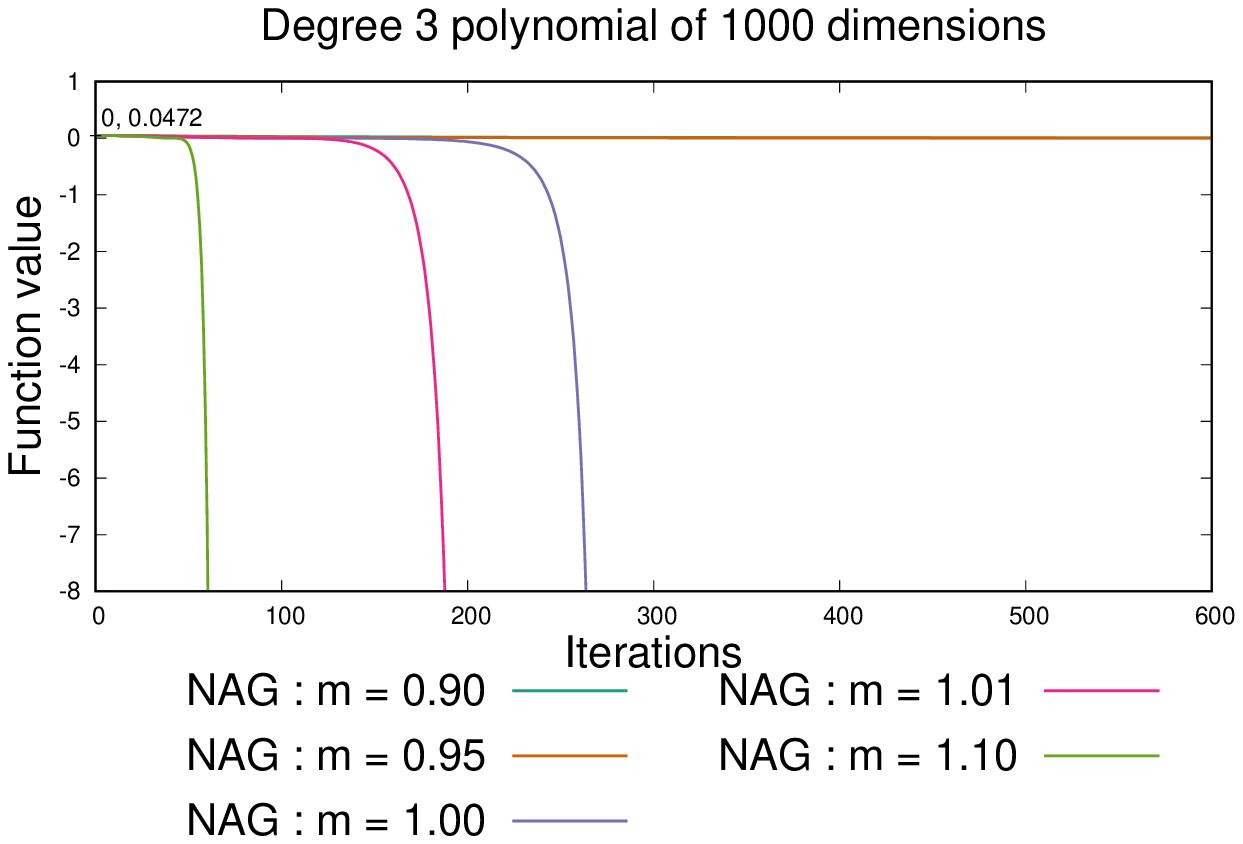}
\end{minipage}

\caption{We set \(n\) to be 100 and 1000 and used \textit{NAG} until the value of the function was lesser than -8. We use 5 different momentum parameters from the set \(\{0.90, 0.95, 1.00, 1.01, 1.10\}\). Note how the descent is faster in the case of \(m \geq 1\) Very similar kinds of results were observed with \textit{CM}. (Best viewed in color)}
\label{fig_synthetic_results-2}
\end{figure}

We also make use of a generalized cubic function of the form 
\begin{equation}
\label{gen-cube}
\displaystyle g(\mathbf{x}) = \mathbf{x}^{T} \Theta (\mathbf{x} \odot \mathbf{x})
\end{equation}
where \(\Theta\) is a diagonal matrix whose entries are sampled from \(\mathcal{U}[1, 2]\), and \(\odot\) represents element-wise product. Note that \(\nabla^{2}g(\mathbf{0}) = \mathbf{0}_{n\times n}\), where \(n\) is the dimensionality of the domain of the function \(g\), meaning \(\mathbf{x}^{*}\) is a saddle-point. The results are shown in Figure \ref{fig_synthetic_results-2}. Note that both the generalized quadratic \ref{gen-quad} and cubic \ref{gen-cube} are unbounded.


\begin{figure}[H]
\begin{minipage}{0.475\linewidth}
\centering
\includegraphics[width=0.95\textwidth]{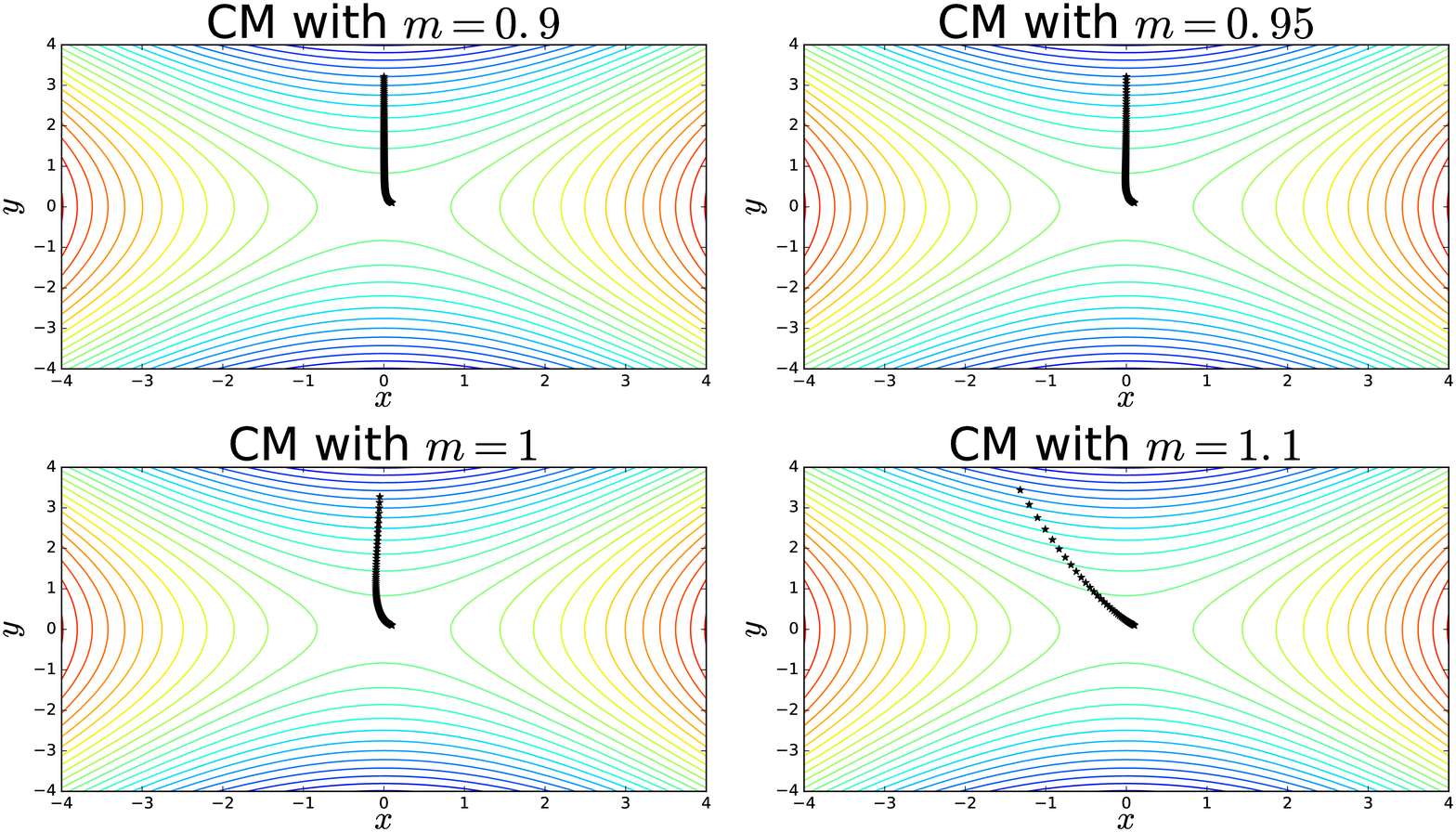}
\end{minipage}
\hfill
\begin{minipage}{0.475\linewidth}
\centering
\includegraphics[width=0.95\textwidth]{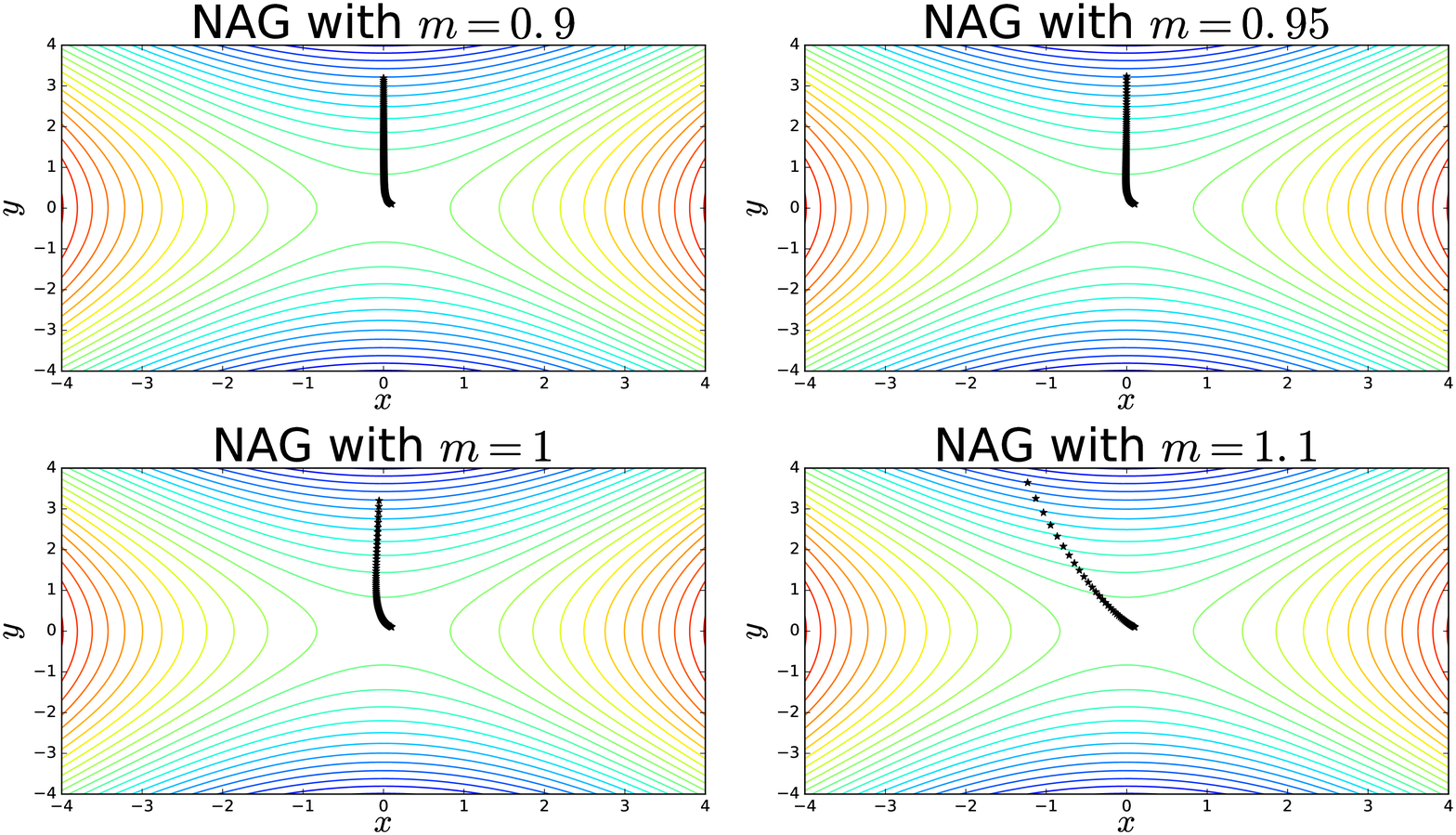}
\end{minipage}

\caption{In both subfigures, the top left contour plot is the trace of the updates obtained by setting m to 0.9, the top right contour plot is the trace of the updates obtained by setting m to 0.95, the bottom left contour plot is the trace of the updates obtained by setting m to 1, and the bottom right contour plot is the trace of the updates obtained by setting m to 1.1.}
\label{2-d-quad}
\end{figure}

Perceiving this idea might be harder due to the existence of higher dimensions, which is why we also provide a contour plot of a special case of the function in equation \ref{gen-quad} with \(n = 2\) and \(\Lambda = \begin{bmatrix} 1 & 0 \\ 0 & -1 \end{bmatrix}\). We try it with both \textit{CM} and \textit{NAG} and wait until the function value dips below -10. The results are tabulated in Table \ref{classic_results} and the traces are shown in Figure \ref{2-d-quad}.

\begin{table}[H]
\caption{Number of iterations taken to reach a function value of -10 with the function \(f(x, y) = x^2 - y^2\) using different momentum schemes.}
\begin{minipage}{0.475\linewidth}
\begin{center}
\begin{tabular}{l|c}
Method & Number of Iterations \\
\hline
\hline
\textit{CM}: m = 0.9 & 209 \\
\hline
\textit{CM}: m = 0.95 & 142 \\
\hline
\textit{CM}: m = 1 & \textbf{93} \\
\hline
\textit{CM}: m = 1.1 & \textbf{49} \\
\hline
\end{tabular}
\end{center}
\end{minipage}
\hfill
\begin{minipage}{0.475\linewidth}
\begin{center}
\begin{tabular}{l|c}
Method & Number of Iterations \\
\hline
\hline
\textit{NAG}: m = 0.9 & 206 \\
\hline
\textit{NAG}: m = 0.95 & 140 \\
\hline
\textit{NAG}: m = 1 & \textbf{91} \\
\hline
\textit{NAG}: m = 1.1 & \textbf{49} \\
\hline
\end{tabular}
\end{center}
\end{minipage}
\label{classic_results}
\end{table}

\subsection{Experiments with Neural Networks}
We also conducted experiments on deep neural networks, in particular, on classification tasks on CIFAR10\footnote{\url{https://www.cs.toronto.edu/~kriz/cifar.html}} and SVHN\footnote{\url{http://ufldl.stanford.edu/housenumbers/}} and autoencoders trained on MNIST\footnote{\url{http://yann.lecun.com/exdb/mnist/}}.

\subsubsection{Experimental setups for SVHN and CIFAR10 classification tasks}
Since our proposed work is an improvement over momentum based methods, we compare our method with the parameterized \textit{NAG} algorithm due to \multiauthor{Sutskever}\cite{Sutskever:2013:IIM:3042817.3043064} with different momentum settings. We use a learning rate of \(10^{-4}\) and a weight decay rate of \(10^{-4}\). The input to the networks are features extracted from the original images using a pre-trained WideResnet \cite{Zagoruyko2016WRN}, which produces 256 and 192 dimensional vectors for each image CIFAR10 and SVHN respectively. 

\subsubsection{Experimental setup for MNIST autoencoder task}
For MNIST, we use the autoencoder architecture described by Hinton and Salakhutdinov in \cite{HinSal06}. We set learning rate and weight decay parameters to be \(10^{-4}\) and \(10^{-4}\) respectively. We also construct a denser version of the aforementioned architecture with an architecture \texttt{784 x 1000 x 500 x 250 x 125 x 60} for the encoder and an architecture \texttt{60 x 125 x 250 x 500 x 1000 x 784} for the decoder. Sigmoid activations are placed between hidden layers of the encoder and decoder, but there is no activation between the encoder and the decoder. We use the same parameters as those used to train the earlier autoencoder.

\subsubsection{Results}
\begin{figure}[H]
\begin{minipage}{0.45\linewidth}
\centering
\includegraphics[width=0.95\textwidth]{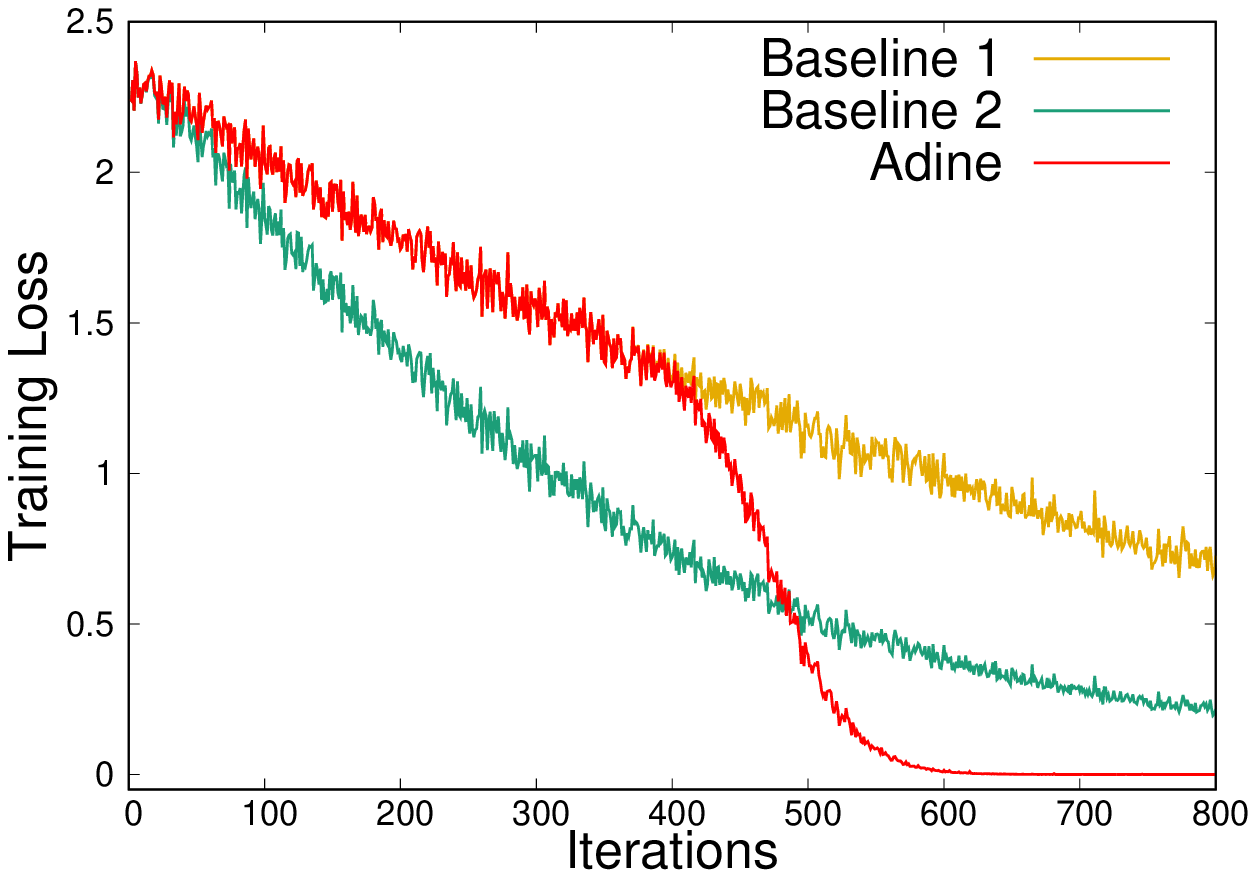}
\end{minipage}
\hfill
\begin{minipage}{0.45\linewidth}
\centering
\includegraphics[width=0.95\textwidth]{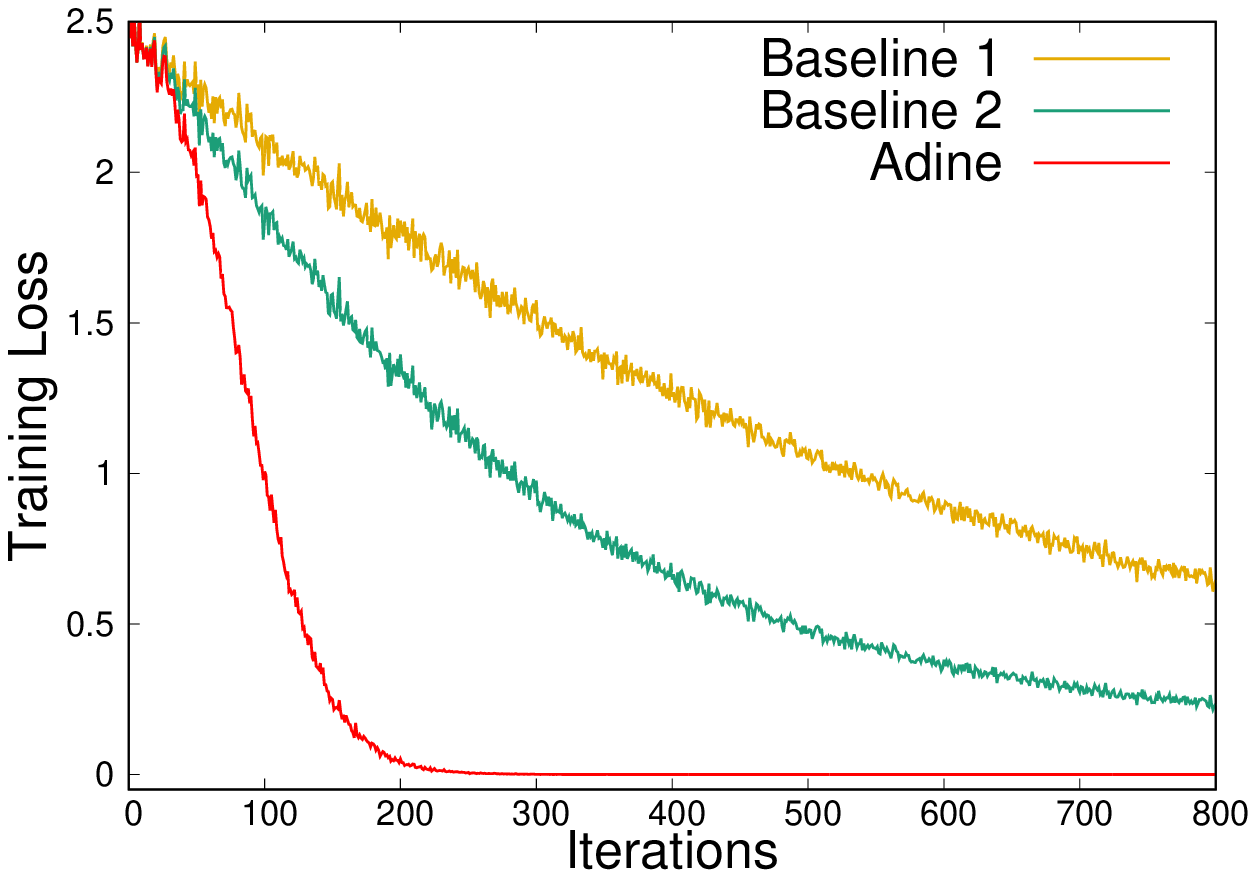}
\end{minipage}
\caption{Convergence of training loss of deep neural network on CIFAR-10 architectures \texttt{256 x 384 x 256 x 128 x 10} (L) and \texttt{256 x 512 x 10} (R). Baseline 1 is \textit{NAG} with m = 0.9, Baseline 2 is \textit{NAG} with m = 0.95, \textit{ADINE} is our proposed method with \(m_s = 0.9\), \(m_{g} = 1.0001\) and \(\zeta = 1.03\) (for L) and \(\zeta = 1.5\) (for R)}
\label{cifar10-res}
\end{figure}

\begin{figure}[H]
\begin{minipage}{0.45\linewidth}
\centering
\includegraphics[width=0.95\textwidth]{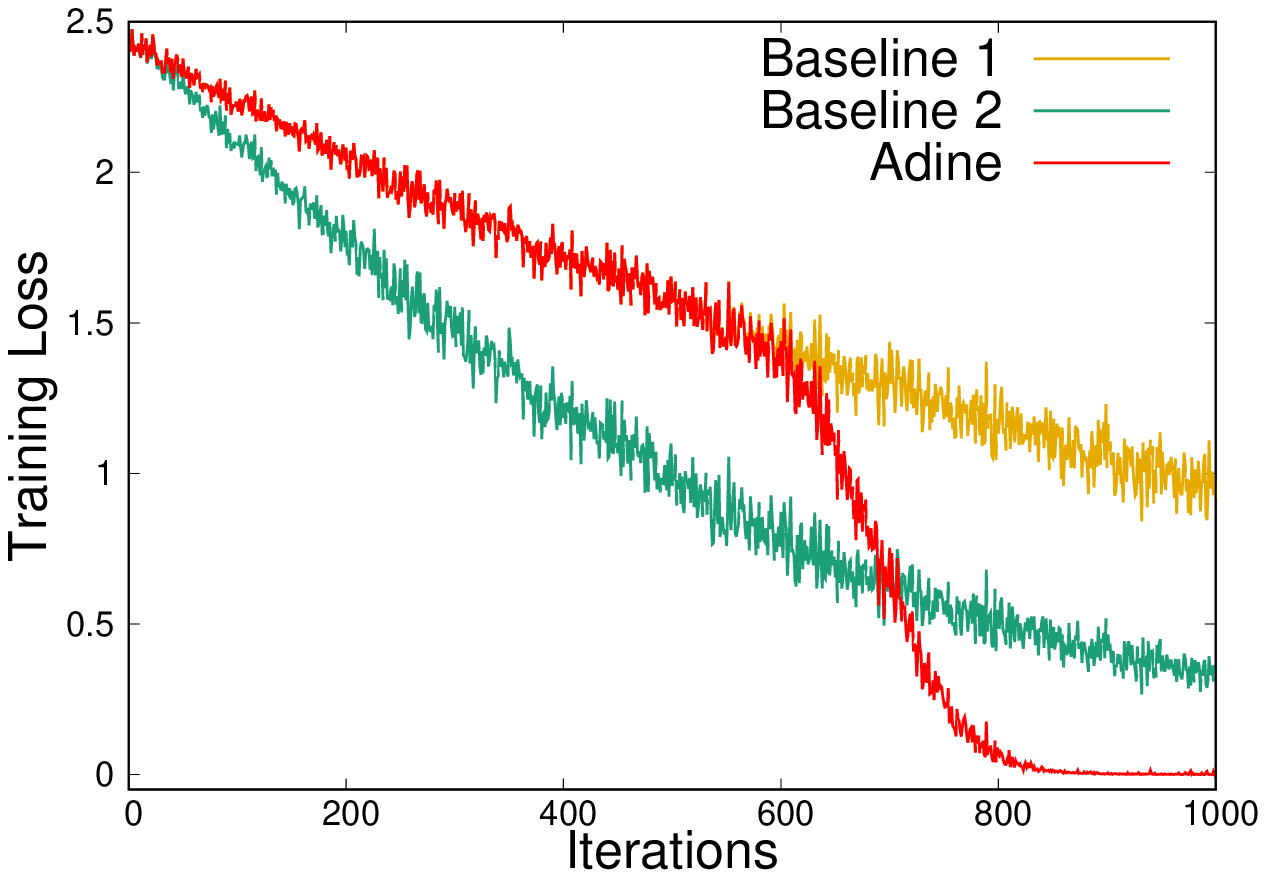}
\end{minipage}
\hfill
\begin{minipage}{0.45\linewidth}
\centering
\includegraphics[width=0.95\textwidth]{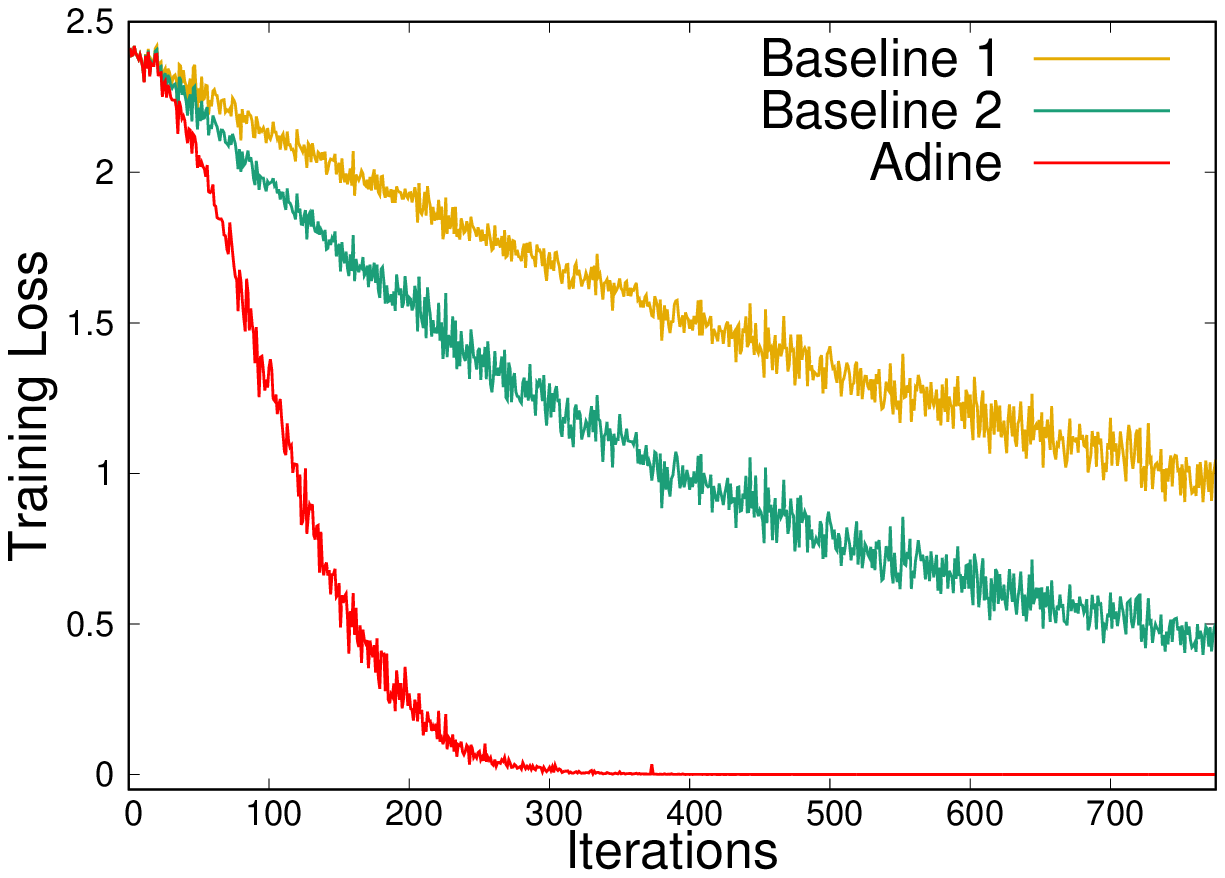}
\end{minipage}
\caption{Convergence of training loss of deep neural network on SVHN architectures \texttt{192 x 288 x 288 x 10} (L) and \texttt{192 x 384 x 10} (R). Baseline 1 is \textit{NAG} with m = 0.9, Baseline 2 is \textit{NAG} with m = 0.95, \textit{ADINE} is our proposed method with \(m_s = 0.9\), \(m_{g} = 1.0001\) and \(\zeta = 1.1\) (for L) and \(\zeta = 1.5\) (for R)}
\label{svhn-res}
\end{figure}

The architectures used for CIFAR10 are \texttt{256 x 384 x 256 x 128 x 10} and \texttt{256 x 512 x 10} and for SVHN are \texttt{192 x 288 x 288 x 10} and \texttt{192 x 384 x 10}. All these architectures have \texttt{ReLU}\cite{Nair:2010:RLU:3104322.3104425} activations in the hidden layers and a \texttt{softmax} activation at the output, and are trained with a cross-entropy loss and a batch size of 32. The weights are initialized with the scheme suggested by Glorot and Bengio \cite{pmlr-v9-glorot10a}. \textit{ADINE} achieves a test accuracy of \(\approx 94\%\) on CIFAR10 and \(\approx 96\%\) on SVHN. These results are consistent with those obtained with standard momentum methods, but \textit{ADINE} is able to achieve this accuracy much faster than the standard momentum methods. Variation of the training loss with time has been plotted in the figures \ref{cifar10-res} and \ref{svhn-res}. 

\begin{figure}[H]
\begin{minipage}{0.45\linewidth}
\centering
\includegraphics[width=0.95\textwidth]{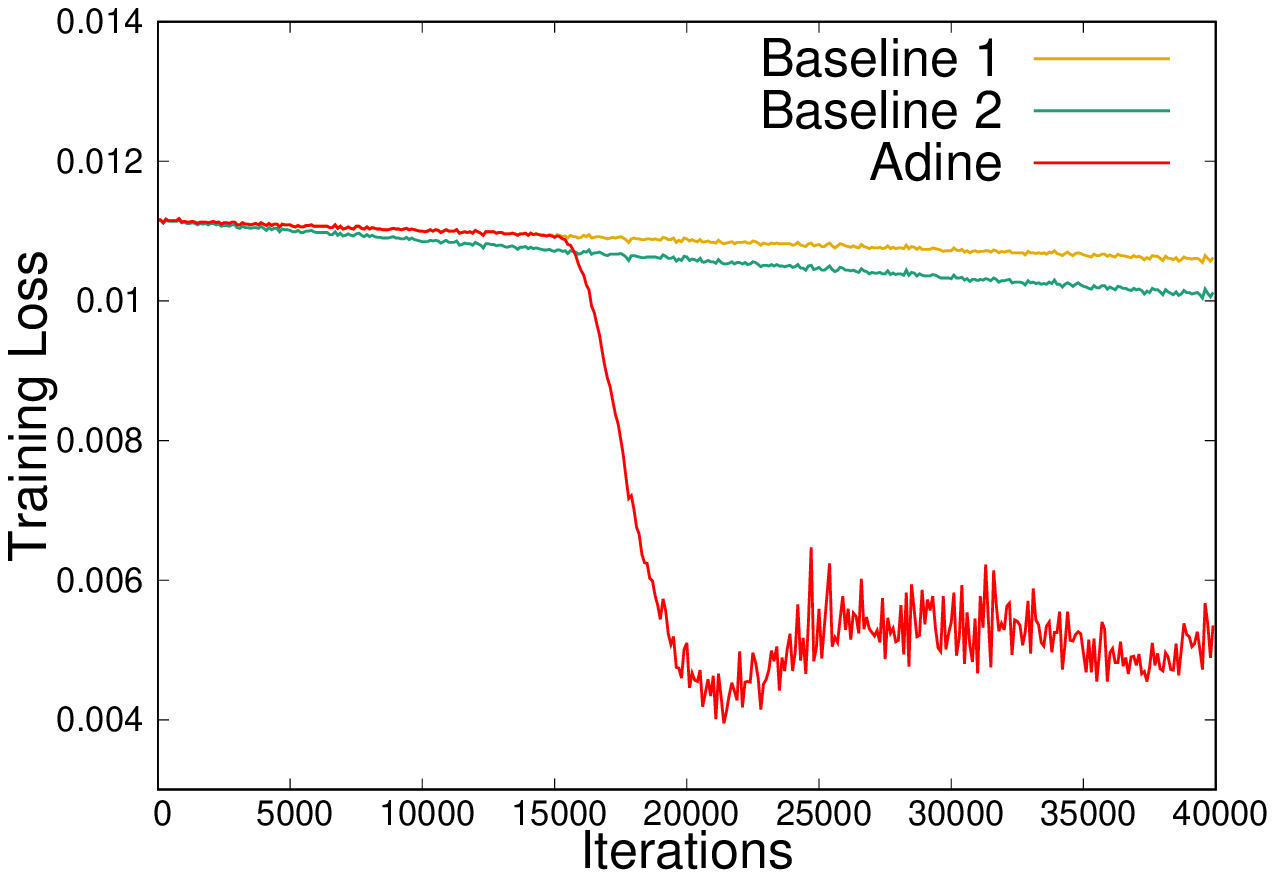}
\end{minipage}
\hfill
\begin{minipage}{0.45\linewidth}
\centering
\includegraphics[width=0.95\textwidth]{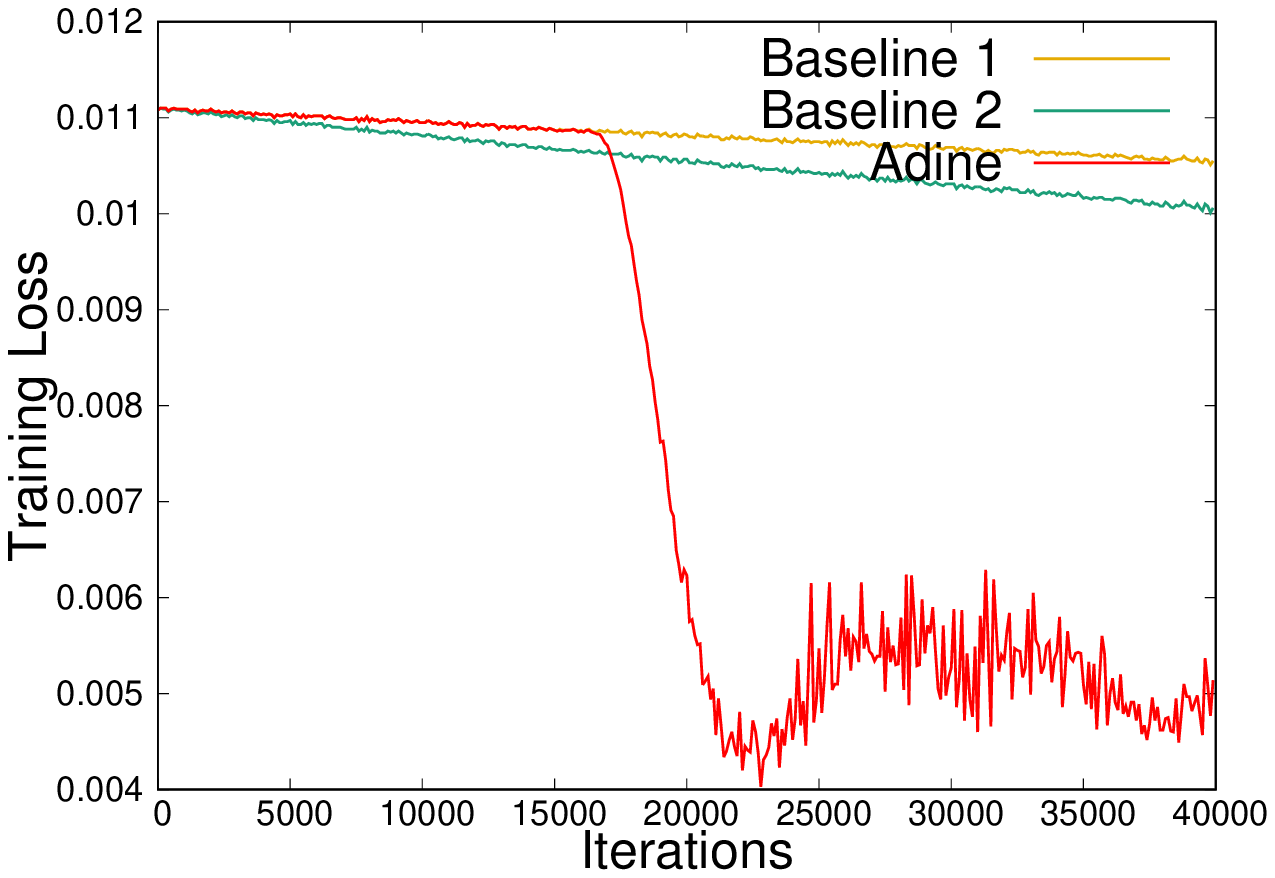}
\end{minipage}
\caption{Convergence of training loss of MNIST autoencoder with architecture due to Hinton and Salakhutdinov \cite{HinSal06} (L) and encoder-decoder architecture of \texttt{784 x 1000 x 500 x 250 x 125 x 60} and \texttt{60 x 125 x 250 x 500 x 1000 x 784} respectively (R). Baseline 1 is \textit{NAG} with m = 0.9, Baseline 2 is \textit{NAG} with m = 0.95, \textit{ADINE} is our proposed method with \(m_s = 0.9\), \(m_{g} = 1.0001\) and \(\zeta = 1.48\) (for L and R)}
\label{mnist-auto}
\end{figure}

For the MNIST autoencoder task, the weights have been initialized using the same scheme as earlier, and is trained with a binary cross entropy loss and a batch size of 64. The variation of the training loss with time has been plotted in Figure \ref{mnist-auto}.

\section{Conclusion}
In this work, we proposed \textit{ADINE}, a faster and way to train SGD using scheduled momentum, where the momentum parameter is adaptively changed based on a new weighted loss sum value in a given iteration. We showed empirically that when training using the momentum scheme, the proposed method, \textit{ADINE}, is able to converge much faster. We demonstrated another major implication of our work in trying escape synthetic saddles in polynomial functions. To the best of our knowledge, this is the first work to explore this particular idea of studying SGD with higher momentum parameter values. As future work, we plan to study the theoretical guarantees of our proposed method \textit{ADINE}.

\section*{Acknowledgements}
We thank Intel India, Microsoft Research India and the Ministry of Human Resource Development, Govt of India for their generous funding for this project. We also thank the developers of \href{https://pytorch.org}{PyTorch} for their work in building the framework for the community.

\bibliographystyle{plain}
\bibliography{main}
\end{document}